# Learning Directed Locomotion in Modular Robots with Evolvable Morphologies

Gongjin Lan[1], Matteo De Carlo[1], Fuda van Diggelen[1], Jakub M. Tomczak[1], Diederik M. Roijers[1,2,3], A.E. Eiben[1]

[1] Department of Computer Science, Vrije Universiteit Amsterdam, The Netherlands
[2] AI laboratory, Vrije Universiteit Brussel, Belgium
[3] Institute of ICT, HU University of Applied Sciences Utrecht, the Netherlands
g.lan@vu.nl, m.decarlo@vu.nl, f.van.diggelen@student.vu.nl,
j.m.tomczak@vu.nl, diederik.yamamoto-roijers@hu.nl, a.e.eiben@vu.nl

**Abstract.** We generalize the well-studied problem of gait learning in modular robots in two dimensions. Firstly, we address locomotion in a given target direction that goes beyond learning a typical undirected gait. Secondly, rather than studying one fixed robot morphology we consider a test suite of different modular robots. This study is based on our interest in evolutionary robot systems where both morphologies and controllers evolve. In such a system, newborn robots have to learn to control their own body that is a random combination of the bodies of the parents. We apply and compare two learning algorithms, Bayesian optimization and HyperNEAT. The results of the experiments in simulation show that both methods successfully learn good controllers, but Bayesian optimization is more effective and efficient. We validate the best learned controllers by constructing three robots from the test suite in the real world and observe their fitness and actual trajectories. The obtained results indicate a reality gap that depends on the controllers and the shape of the robots, but overall the trajectories are adequate and follow the target directions successfully.

## 1 Introduction

Developing robots for known and structured environments is a challenging task, however, it is considerably harder for (partially) unknown and complex environments, like deep seas, rain forests or other planets. The challenge here is twofold: the unpredictability and the complexity of the environment. Designing appropriate robot morphologies and corresponding controllers for such environments is a daunting task where classic engineering approaches seem to fall short. Therefore, it would be highly useful to use an evolutionary approach for such problems, where robots evolve their morphologies and controllers over generations to better adapt to the environment.

The field that is concerned with such evolving robots is Evolutionary Robotics [10,19,38]. To date, this research community has mainly been focusing on evolving only the controllers in fixed robot bodies. The evolution of morphologies has received much less attention, even though it has been observed that adequate robot behaviour depends on both the body and the brain (controller) [5,7,39]. To unlock the full potential of the evolutionary approach, especially for unknown and/or changing environments, one should apply it to both bodies and brains and even to the materials used in the robot components [25].

A generic architecture of robot systems, where both morphologies and controllers evolve in real time and real space has been introduced in [20,21], and a proof-of-concept has been conducted recently [29]. The underlying model, called the Triangle of Life (ToL), describes a life cycle that runs from conception (being conceived) to conception (conceiving offspring) through three principal stages: Birth, Infancy, and Mature Life as illustrated in Figure 1.

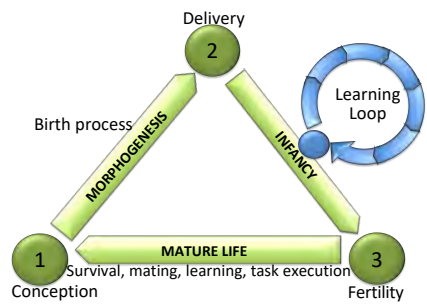

Fig. 1: Generic system architecture represented by the Triangle of Life. The pivotal moments that span the triangle and separate the three stages are: 1) Conception: A new genotype is activated, construction of a new robot starts. 2) Delivery: Construction of the new robot (the phenotype) is completed. 3) Fertility: The robot becomes an adult, ready to conceive offspring.

One of the challenges inherent to evolving robot bodies – be it simulated or real – is rooted in the fact that 'robot children' are random combinations of the bodies and brains of their parents. In general, it cannot be assumed that simply recombining the parents' controllers results in a controller that fits the recombined body. Hence, a 'robot child' must undergo a learning process to learn to control its body, not unlike a little calf that spends the first hour of its life learning to walk. It is vital that the learning method is general enough to work for a large variety of morphologies, and fast enough to work within practical time intervals.

In this paper, we address a special case of this generic problem, that of acquiring directed locomotion skills. Formerly, we have investigated the most elementary case, gait learning [29,30,31,46]. However, although gait learning is a popular problem within evolutionary robotics, in practice we are typically not only interested in a robot that just walks without purpose. For most cases, a robot has to move in a target direction, e.g., to move towards a destination or to systematically explore its environment. Therefore, we focus on the task of directed locomotion, where the robot must follow a given direction, e.g. "go

North" or "go forward". To validate our results, we test the best controllers on real robots as well. The corresponding general problem statement is as follows:

> How to learn controllers efficiently for directed locomotion in a variety of modular robots with different morphologies?

The paper addresses this problem by three specific research goals:

1. Describe possible candidates, i.e., algorithms to learn controllers for directed locomotion in modular robots.
2. Evaluate these algorithms on a test suite of modular robots with different shapes and sizes by comparing their efficacy and efficiency in simulation.
3. Demonstrate the learned directed locomotion behavior on real robots and reflect on the reality gap in this application.

## 2 Related work

Evolutionary robotics aims to design adaptive robots automatically that can evolve to accomplish a specific task while adapting to environmental changes [2]. A number of studies have demonstrated the feasibility of evolutionary methods for generating robotic control and/or morphology [29].

The design of locomotion for modular robots with evolvable morphologies is a difficult task. Several approaches based on various types of controllers and algorithms for locomotion of robots have been proposed in [3,42]. An early approach is based on gait control tables that, in essence, are a simple cyclic finite state machine [9]. A second major approach is based on neural networks, for instance, HyperNEAT [43]. In previous work [23,46], we have implemented evolutionary controllers for locomotion in modular robots using HyperNEAT. Other studies [16,49] have also shown that HyperNEAT can evolve good controllers for efficient gaits of robots. Moreover, Bayesian optimization have been proven to perform the state-of-the-art data-efficient learning of controllers for robots [17]. [13,12] present efficient gait learning on a physical bipedal robot within limited trials by Bayesian optimization. In recent studies [35,48], Bayesian optimization achieves data-efficient learning of locomotion on a 6-legged microrobot. Other successful approaches that have been extensively investigated for robot locomotion are based on Central Pattern Generators (CPG) [27]. CPGs are neural networks that can produce rhythmic patterned outputs without rhythmic sensory or central input [24]. The CPG-based controller allows one to reduce the dimensionality of the locomotion control problem while remaining highly flexible to continuously adjust velocity, direction, and type of gait according to the environmental context [28]. This technique has been shown to produce well-performing and stable gaits for modular robots [32,33,36]. Lastly, an alternative approach based on machine learning for adaptive locomotion was proposed by Cully et al. [18].

Although there are extensive studies on the locomotion of robots, most of them focus on the controllers in fixed robot bodies for gait learning, and only some of them consider multiple shapes of robots [32,42]. Voxel soft and modular robots are studied with evolvable morphologies in simulation for gait learning in [15,45] respectively. Dylan et al. [6] present the jointly evolution of both controllers and bodies, while it studies on the locomotion task in simulation with simple ant robots. Ahmadzadeh and Masehian [1] reviewed many fixed special modular robots for the tasks of gait learning, self-reconfiguration, self-assembly, self-disassembly, self-adaptation, and grasping in simulation or/and real-world. However, none of them achieve directed locomotion in modular robots with evolvable morpholoiges. In this paper, we aim to evolve modular robots that can move towards target directions.

Most existing studies in directed locomotion for the robots are about the control of vertebrates with fixed shapes, such as a biped [22]. There were many neural-based control systems for directed vertebrates locomotion proposed. For instance, a CPG approach based on phase oscillators towards directed biped locomotion is outlined in [37]. In [14], a reinforcement learning is used to learn directed locomotion for a special snake-like robot with screw-drive units. However, there are only few studies that investigate directed locomotion of modular robots, and they focus on fixed morphologies or special structures.

Even though evolutionary robotics is not new, rarely it has been used to generate a physical demonstration of evolved robots, due to the wear and tear of hardware, the infamous reality gap, etc. In [40], it is estimated that more than 95% of the literature is focused on the evolution of robot controllers in simulation, and less than 1% of reported works physically tested the generated morphologies and controllers in the remaining 5%. Among them, Sproewitz et al. [42] implemented three modular robots with fixed morphology for gait learning with CPG controller in a simulated environment and real-world. In other studies, a gait learning task in the real world are performed on a quadruped robot [28], salamander robot[44], and modular robot [49]. Although [9] studied the directed locomotion in simulation and real-world, they only focused on the task in a quadruped robot without multiple and evolvable morphologies. [37] studied on a fixed biped robot in simulation and real world for directed locomotion. Even though [32,33] investigated the modular robots with multiple morphologies in both simulation and real-world, they mainly focused on gait learning. In this paper, we are concerned with learning controllers for directed locomotion on modular robots with evolvable morphologies in simulation and demonstrate the best controllers in real-world. We show an overview of the related work and the position of our work in Table 1.

| Articles | Robots | Morphology | Environments | Tasks | Year |
|---|---|---|---|---|---|
| [32,33] | Modular robots | Multiple fixed | SIM & **RW** | GL | [2005,2004] |
| [9] | Quadruped robot | One fixed | SIM & **RW** | **DL** | [2006] |
| [28] | Salamander robot | One fixed | **RW** | **DL** | [2007] |
| [42] | Modular robots | Three fixed | SIM & **RW** | GL | [2008] |
| [16] | Quadruped robot | One fixed | SIM | GL | [2009] |
| [23] | Modular robots | One fixed | SIM | GL | [2010] |
| [49] | Quadruped robot | One fixed | **RW** | GL | [2011] |
| [37] | Biped robot | One fixed | SIM & **RW** | **DL** | [2014] |
| [44] | Modular robot | One fixed | **RW** | GL | [2014] |
| [8] | Modular robots | **Evolvable** | SIM | GL, others | [2014] |
| [29,30,45,46] | Modular robots | **Evolvable** | SIM | GL | [2017,16,17,17] |
| [1] | Modular robots | Multiple fixed | SIM & **RW** | GL, others | [2017] |
| [15] | Voxel soft robots | **Evolvable** | SIM | GL | [2018] |
| [6] | Modular ant robots | **Evolvable** | SIM | GL | [2019] |
| **Our work** | **Modular robots** | **Evolvable** | **SIM & RW** | **DL** | - |

Table 1: Overview of related work and the position of our work. SIM: simulation; RW: real-world; GL: gait learning; DL: directed locomotion.

## 3 Robot system

### 3.1 Robot morphologies

We design the modular robots using a subset of the components: *brick components*, a *core component*, and *active hinges* [29], which is based on Robogen [4]. The brick components are cubic with four slots on their sides available for other modules to attach to. The core component is a larger *brick* that holds the controller board and a battery. It also has four slots on its lateral faces to attach other components. The active hinge component is a joint actuated by a servo motor. It can attach other components on two opposite sides by inserting its lateral faces into the slots of other components. Each robot's genotype describes its layout and consists of a tree structure with the root node representing a core module from which further components branch out. Component types contain specific features described by its genotypical encoding dependant on a component's type. These models are used in the simulation, but could also be used for 3D printing and construction of the real robots.

As a test suite we chose nine modular robots in three different shapes and sizes that represent "parent" and "baby" robots. We refer to the three shapes as *spider*, *gecko*, and *baby* (see Figure 2). The "baby" robots were created through recombination or mutation of the "spider" and "gecko" [30]. Each type of shape includes three sizes: small (7, 8 or 9 bricks for, *spider*, *gecko*, and *baby*, respectively), medium (11, 12 or 13 bricks), and large (15 or 17 bricks), resulting in the nine robots shown in Figure 2.

### 3.2 Robot controllers

CPG-based Controllers have been proven to perform well in modular robotics [27]. A CPG controller is a neural circuit, in which the activation functions of neurons is flexible. While different types of CPGs exist as a robot controller, the CPG with a system of Ordinary Differential Equations (ODEs) as activation function is often used, i.e., differential CPG [47]. As a result, the output values of the nodes are subject to oscillatory behaviour.

In this work, we implement the CPGs whose main components are differential oscillators. Each robot joint has a differential oscillator that is defined by two neurons, a $x_i$-neuron, a $y_i$-neuron, and $out_i$-neuron that are recursively connected as shown in Figure 3. The index $i$ represents the number of a differen-tial oscillator, and $w_{x_i y_i}$, $w_{y_i x_i}$, and $w_{x_i o_i}$ denote the weights of the connections between the neurons. The $x_i$-neuron and $y_i$-neuron feed their activation values multiplied by weights $w_{x_i y_i}$ and $w_{y_i x_i}$ to the $y_i$-neuron and $x_i$-neuron respectively. At a time step $t$, the changes of activation value of $x_i$-neuron ($dx_{(i,t)}$) and $y_i$-neuron ($dy_{(i,t)}$) can be calculated according to the following differential equation:

$$\begin{aligned} \frac{dx_{(i,t)}}{dt} &= w_{y_i x_i} y_{(i,t-1)} \\ \frac{dy_{(i,t)}}{dt} &= w_{x_i y_i} x_{(i,t-1)} \end{aligned} \quad (1)$$

where $t-1$ represents the last time step. $x_i$-neuron and $y_i$-neuron generate the activation values $x_{(i,t)}$ and $y_{(i,t)}$ of oscillatory patterns over time according to the following expression:

$$
\begin{aligned}
x_{(i,t)} &= x_{(i,t-1)} + \frac{dx_{(i,t)}}{dt} \\
y_{(i,t)} &= y_{(i,t-1)} + \frac{dy_{(i,t)}}{dt}
\end{aligned}
\quad (2)
$$

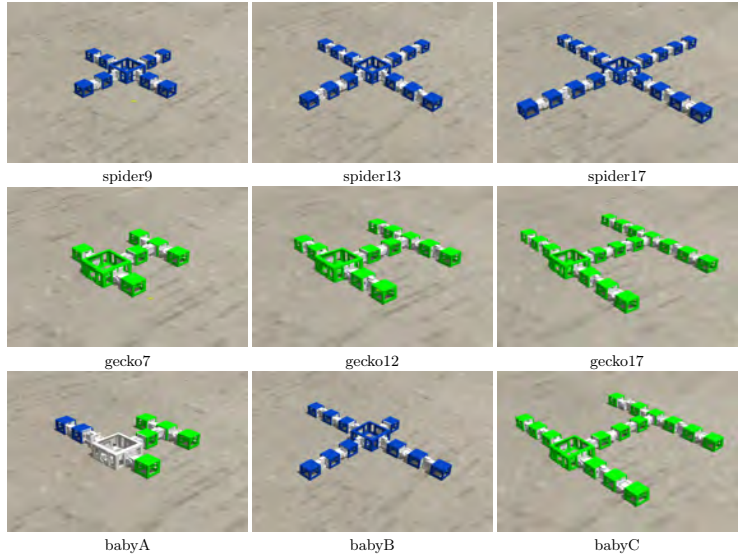

Fig. 2: Images of the simulated modular robots. The top and middle rows exhibit the basic shapes named spider, gecko in three sizes separately. The bottom row shows the three 'baby' morphologies created through recombining or/and mutating basic shapes. Please note that the top leg of gecko17 and that of babyC are different; specifically, babyC has one more active hinge in the top leg, where gecko17 has a brick and an active hinge.

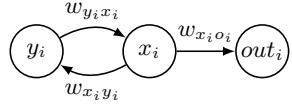

Fig. 3: A differential oscillator $i$ with output node in a joint of modular robots as used in the CPG controller.

The $x_i$-neuron feeds its activation value multiplied by the weight $w_{x_i o_i}$ to the $out_i$-neuron, the $out_i$-neuron applies the activation function and generates the driving signal to the servo in a joint of the robot. As the differential oscillator in a robot controller, the activation values of the neurons have to meet two conditions. First, the activation value of $out_i$-neuron should be bounded due to the limited rotating angle of the joints. Therefore, we use a variant of the sigmoid function, the hyperbolic tangent function ($tanh$), as the activation function of $out_i$-neurons to bound the output value in $[-1, 1]$. At a time step $t$, the $tanh$ activation value of $out_i$-neuron can be calculated as follows:

$$
out_{(i,t)}(x_{(i,t)}) = \frac{2}{1 + e^{-2x_{(i,t)}}} - 1 \quad (3)
$$

Second, the activation value of $out_i$-neuron should be periodic. The weights with the same values but different signs, i.e., $w_{x_i y_i} = -w_{y_i x_i}$, achieve periodic activation values of neurons. We use the predefined initial values $(x_{(i,0)}, y_{(i,0)}) = (-\frac{1}{2}\sqrt{2}, \frac{1}{2}\sqrt{2})$, and $(w_{x_i y_i}, w_{y_i x_i}) = (0.5, -0.5)$, but they can be randomly initialized except 0. In such a way, the differential oscillator can generate the oscillatory activation values.

An independent differential oscillator generates only sinusoidal waves. However, for the modular robot controllers, we implement the CPG controllers with the connections of the neighbouring differential oscillators. For instance, for the specific morphology of the modular robot spider9, its CPG network has connections between the neighboring differential oscillators, as shown in Figure 4. As a result, the output values are a composition of multiple sinusoidal signals.

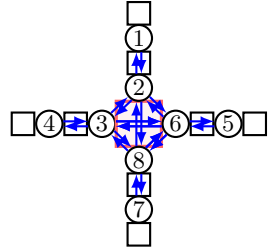

Fig. 4: Schematic view of CPG network generated for the robot *spider9*. The squares represent the components of fixed bricks. The circles with numbers represent the differential oscillators in the joints (active hinges).

Combining the differential oscillator (rf. Figure 3) into each joint (the circle with number), the CPG network of the robot spider9 can be specified as shown in

$$x_{(i,t)} = x_{(i,t-1)} + \frac{dx_{(i,t)}}{dt} + \sum_{j \in \mathcal{N}_i} x_{(j,t-1)} w_{ji}$$
$$y_{(i,t)} = y_{(i,t-1)} + \frac{dy_{(i,t)}}{dt} \quad (4)$$

where $i$ is the number of the differential oscillator, $\mathcal{N}_i$ is the set of indices of the neighbouring differential oscillators. For instance, the differential oscillators 1 and 2 have $\mathcal{N}_1 = \{2\}$ and $\mathcal{N}_2 = \{1, 3, 6, 8\}$, respectively, for the CPG network of the robot spider9. Subsequently, the output value of each differential oscillator can be calculated by Equation 3. To reduce the number of weights to be learned, we set $w_{x_i o_i} = 1.0$ in this work, i.e., the input of the $out_i$-neuron equals the activation value $x_{(i,t)}$ of the $x_i$-neuron. Furthermore, we apply $w_{ji} = -w_{ij}$ to the CPG controllers.

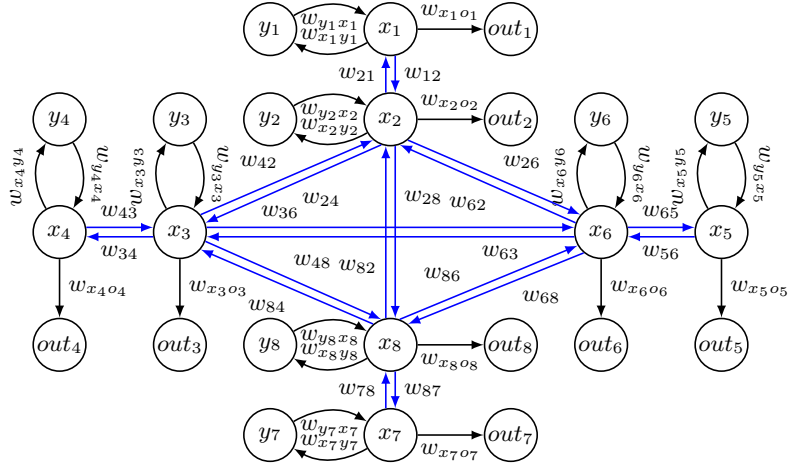

Fig. 5: The specific network of CPG controller generated for a specific morphology of the modular robot spider9.

Figure 5. With the connections to the neighbouring differential oscillators, the activation values of each differential oscillator can be calculated as follows:

To be able to optimize CPG controllers with Bayesian optimization and HyperNEAT, we need a unique one-to-one topology mapping for each weight. In our system, each node of a CPG controller has an unique three-dimensional coordinate that is determined in two steps. First, a differential oscillator of the CPG is encoded to a two dimensional coordinate that corresponds to the relative position of the given active hinge. Second, the third coordinate is defined depending on the nodes: output nodes ($out_i$-neurons) are given a value of 0, while differential nodes are given the value of 1 for $x_i$ the node ($x_i$-neuron) and $-1$ for the $y_i$ node ($y_i$-neuron).

Moreover, we need to specify source nodes and target nodes for distinguishing the weights in CPG controllers. Therefore, we combine the three dimensional coordinates of two nodes as a unique six dimensional information of a connection weight from a source node to a target node. For instance, if $(a, b)$ is the two dimensional coordinate of a differential oscillator, then the weight $w_{y_i x_i}$ from the $y_i$-neuron to the $x_i$-neuron has the six dimensional vector $(a, b, -1, a, b, 1)$. Similarly, $(a, b, 1, a, b, -1)$ corresponds to the weight $w_{x_i y_i}$. The number of the parameters that need to be optimized for the nine modular robots are shown in Table 2. Recall that we apply $w_{ij} = -w_{ji}$, and $w_{x_i o_i} = 1.0$ to the CPG controllers for reduction of the number of parameters to be learned and $w_{x_i y_i} = -w_{y_i x_i}$ for the periodicity of the activation value of neurons. Optimizing the weights of the CPG controllers can achieve expected behaviours of modular robots with different morphologies. Therefore, the outstanding optimization algorithms (learners) are crucial to search the optimal CPG controllers by optimizing the weights.

| Robots | Spdier9 | Spider13 | Spider17 | Gecko7 | Gecko12 | Gecko17 | BabyA | babyB | babyC |
|---|---|---|---|---|---|---|---|---|---|
| Weights | 18 | 26 | 34 | 13 | 23 | 33 | 16 | 22 | 32 |

Table 2: Number of parameters in the controllers of the robots in our test suite.

### 3.3 Learning algorithms

We apply two algorithms, Bayesian optimization and HyperNEAT, to learn the weights of CPG controllers for the test suite of nine modular robots with different shapes and sizes. For each robot we test the learning algorithms on five target directions ($40°$, $20°$, $0°$, $-20°$, and $-40°$ relative to the robot) to simulate the robot's limited field of view in the real-world. The CPG controller of each modular robot with certain topology is automatically generated according to its genotypical encoding. The learning run is repeated ten times for each robot and each target direction. The overall architecture of the learning system is shown in Figure 6, including three stages:

1. Generating topologies of CPG controllers automatically according to the genotypical encoding of different modular robots. That is shown as the blue blocks on the left of the figure. For any certain robot, the topology of CPG controller is fixed as it is fully determined by the given morphology.
2. Learning weights of CPG controllers by learning algorithms. The algorithms learn the weights of CPG controllers that their topologies are generated in the first stage. We apply Bayesian optimization and HyperNEAT to be the learners, as shown in the two red blocks. The algorithms take the information of CPG topologies and output the learned value of the corresponding weights.
3. Evaluating the behaviours of modular robots with the learned controllers by the fitness function. The CPG controllers with the same topology but different value of the weights drive the modular robot to perform different behaviors, that in turn obtains different fitnesses.

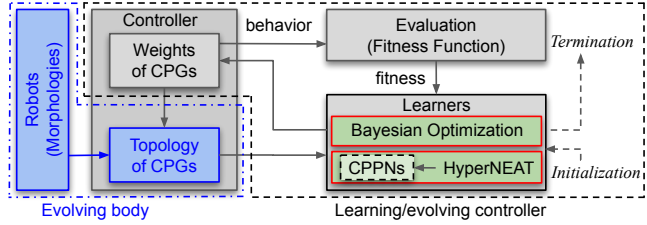

Fig. 6: The overall architecture of the learning system. The learning methods are implemented by Bayesian optimization and HyperNEAT. The learners optimize the weights of the CPG controllers whose topologies are generated automatically according to the (evolved) morphologies of modular robots.

In the grand scheme of the Triangle of Life, in the context of infancy, the learning loop is the combination of the second and the third stage. A CPG topology of a modular robot is generated in the first stage, subsequently the learning loop consisting of the second and third stage are executed for learning the optimal weights of the CPG controller until a termination condition is triggered, i.e., 1500 evaluations. That is, for the first stage, there are 1500 combinations of the second and third stage executed and 10 repetitions. The fitness reflects the quality of the learned weights that the learners aim to maximize.

**Bayesian Optimization** Bayesian optimization is a state-of-the-art machine learning framework for optimizing stochastic functions, and has been successfully applied in engineering, machine learning, and design [41]. It aims to find an optimal solution in a small number of function evaluations that is especially advantageous in situations where evaluations are costly [11]. This can be achieved by constructing the surrogate model using a Gaussian process (GP) approximation of the objective function of a task, and selecting a new controller with the weights $\overrightarrow{\mathcal{W}}_k(w_1, w_2, ..., w_j)$ to evaluate on the basis of an acquisition function that balances exploration and exploitation. The surrogate model is updated with the evaluation of a new controller. Then, the learning system selects the next new controller and repeats the learning process until it reaches the termination criterion. The performance of a CPG controller is evaluated in a given evaluation time by the fitness function defined below. We empirically set the evaluation time to be 60 seconds to balance computing time and accurately evaluating a complex task as directed locomotion. The pseudocode of Bayesian optimization for optimizing the weights of CPG controllers is shown in Algorithm 1.

In this paper, we use the upper confidence bound (UCB) as the acquisition function and Matérn 5/2 kernel as the kernel function. Particularly, we use Latin Hypercube Sampling (LHS) to generate 50 initial samples of the weights instead of random sampling that Bayesian optimization typically uses. In our preliminary experiments, Bayesian optimization with LHS performs better than with random sampling for all nine modular robots. The main parameters of Bayesian

---

**Algorithm 1** Bayesian optimization for learning the weights of CPG controllers.

---
1: generate $n$ initial controllers $\vec{\mathcal{W}}_1, \vec{\mathcal{W}}_2, ..., \vec{\mathcal{W}}_i, ..., \vec{\mathcal{W}}_n$; $n = 50$ in this work, $\vec{\mathcal{W}}_i$ is a $j$ dimensional vector $\vec{\mathcal{W}}_i(w_1, w_2, ..., w_j)$ that is the weights extracted automatically according to the morphologies of different modular robots.
2: evaluate the initial controllers by the fitness function in Equation 9 to obtain fitness $f_1, f_2, ..., f_i, ..., f_n$.
3: get the initial GP: $\mu(\vec{\mathcal{W}}_{1:n})$, $\sigma^2(\vec{\mathcal{W}}_{1:n})$.
4: **for** $k = n+1, n+2, ...$ **do** ▷ k is the index of evaluations
5:     select a new controller $\vec{\mathcal{W}}_k$ by optimizing acquisition function $u(\vec{\mathcal{W}}_k)$:

$$\vec{\mathcal{W}}_k = \arg\max_{\vec{\mathcal{W}}_k} u(\vec{\mathcal{W}}_k | \vec{\mathcal{W}}_{1:k-1})$$

6:     evaluate the new controller $\vec{\mathcal{W}}_k$ to obtain the fitness $f_k$.
7:     augment data $(\vec{\mathcal{W}}_{1:k}, f_{1:k}) = \{\vec{\mathcal{W}}_{1:k-1}, (\vec{\mathcal{W}}_k, f_k)\}$.
8:     update GP: $\mu_k(\vec{\mathcal{W}}_{1:k})$, $\sigma_k^2(\vec{\mathcal{W}}_{1:k})$.
9: **end for**
10: **return** data $(\vec{\mathcal{W}}_{1:k}, f_{1:k})$.

---

optimization and the tuned values that we used for our experiments, are shown in Table 3.

| Parameters | Value | Description |
|---|---|---|
| Initial samples | 50 | The number of initial samples. |
| learning iterations | 1450 | The number of evaluations, excluding initial samples. |
| Kernel variance | 1.0 | The kernel variance in Matérn 5/2 kernel. |
| Kernel length | 0.2 | The characteristic length-scale in Matérn 5/2 kernel. |
| UCB alpha | 3.0 | The weight in the acquisition function. |
| Initial sampling | LHS | The method used to generate initial sampling. |

Table 3: Main experimental parameters of Bayesian optimization.

**HyperNEAT** It has been repeatedly demonstrated that HyperNEAT performs well to learn controllers in modular robots for a given task [23,34,31,49]. HyperNEAT generates Compositional Pattern-Producing Networks (CPPNs) that are a variation of artificial neural networks (ANNs) but evolving. The unique six dimensional information (Section 3.2) of a weight in CPG controllers is the input of the evolved CPPN. The CPPN outputs the values of the weights that in turn constitute the CPG controller that induces the behaviour for directed locomotion. The behaviour is evaluated by a fitness function (Section 3.4) and the fitness value is returned to HyperNEAT which in turn generates new CPPNs, closing the loop (Fig. 6). CPPNs evolve until a termination condition is triggered; in our experiments this is reaching a maximum number of generations. The pseudocode of HyperNEAT for generating CPPNs to optimize the weights of CPG controllers is shown in Algorithm 2.

---

**Algorithm 2** HyperNEAT learns the CPG controllers.
---
1: generate $n$ initial CPPNs, noted $\mathcal{N}_1, \mathcal{N}_2, ..., \mathcal{N}_i, ..., \mathcal{N}_n$, population size $n = 20$.
2: evaluate the CPG controllers $\overrightarrow{\mathcal{W}}_1, \overrightarrow{\mathcal{W}}_2, ..., \overrightarrow{\mathcal{W}}_i, ..., \overrightarrow{\mathcal{W}}_n$ generated by the initial CPPNs to obtain the fitness $f_1, f_2, ..., f_i, ..., f_n$.
3: **for** $k = 2, 3, ...$ **do** ▷ $k$ is the index of generation
4:     generate new candidate CPPNs, $\mathcal{N}'_{((k-1)*n+1):(k*n)}$ by mutation and crossover.
5:     evaluate the CPG controllers $\overrightarrow{\mathcal{W}}'_{((k-1)*n+1):(k*n)}$ generated by the CPPNs $\mathcal{N}'_{((k-1)*n+1):(k*n)}$ to obtain the fitness $f'_{((k-1)*n+1):(k*n)}$.
6:     select the CPPNs from $\mathcal{N}'_{((k-1)*n+1):(k*n)}$ by the fitness to be the CPPNs $\mathcal{N}_{((k-1)*n+1):(k*n)}$ in next generation, and update the corresponding CPG controllers $\mathcal{W}_{((k-1)*n+1):(k*n)}$ and their fitness $f_{((k-1)*n+1):(k*n)}$.
7:     update $\overrightarrow{\mathcal{W}}_{1:(k*n)}$ and their fitness $f_{1:(k*n)}$.
8: **end for**
9: **return** data $(\overrightarrow{\mathcal{W}}_{1:(k*n)}, f_{1:(k*n)})$.
---

An initial population of 20 CPPNs are randomly generated in the first generation. Each CPPN generates weights of a CPG controller whose topology is based on a robot's morphology. The performance of the CPG controller is evaluated in 60 seconds as the same duration with Bayesian optimization. Each learning run is terminated after 75 generations with 20 populations, that is, 1500 fitness evaluations. The experimental parameters we used in the experiments are described in Table 4.

| Parameters | Value | Description |
| --- | --- | --- |
| Mutation | 0.8 | Probability of mutation for individuals |
| Generations | 75 | Termination condition for each run |
| Population size | 20 | Number of individuals per generation |
| Tournament size | 4 | Number of individuals used in tournament selection |

Table 4: Main experimental parameters of HyperNEAT

### 3.4 Fitness Function for Directed Locomotion

Learning CPG controllers for directed locomotion in modular robots with evolvable morphologies is a black-box optimization problem, we therefore need to formulate a fitness function to the objective. In our system, fitness function is not only used to evaluate the performance of controllers but also serves as the guiding metric of learning controllers. We define a fitness function for directed locomotion that combines two objectives: minimizing deviation with respect to the target direction and maximizing speed with minimum length of the trajectory. In this section, we provide a step-by-step derivation, culminating in the final fitness function stated in Equation 9.

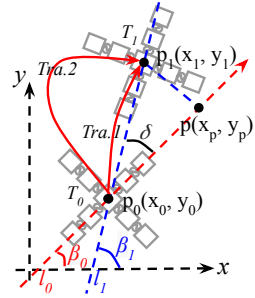

Fig. 7: Illustration of the fitness calculation. $p_0(x_0, y_0)$ is the starting position of the robot, $p_1(x_1, y_1)$ is the end position. The red line $l_0$ shows the target direction, the blue line $l_1$ is the direction actually travelled between $p_0$ and $p_1$. The angle $\delta$ is the deviation between $l_0$ and $l_1$. The point $p(x_p, y_p)$ is the projection of $p_1$ on the line $l_0$. The red lines $Tra.1$ and $Tra.2$ show two different trajectories between $p_0$ and $p_1$.

The scenario for an evaluation in our experiments is illustrated in Fig. 7. We can collect the following measurements from the Revolve framework:

1. $p_0(x_0, y_0)$ is the coordinate of the core component of the robot at the start of the simulation, i.e., time $T_0$.
2. $p_1(x_1, y_1)$ is the coordinate of the core component of the robot at the end of the simulation,i.e., time $T_1$.
3. The orientation of the robot in time $T_0$.

The target direction, $\beta_0$, is an angle with respect to the initial orientation of the robot at time $T_0$. In Fig. 7 we drew lines in the target direction, $l_0$, and the line $l_1$ through $p_0$ and $p_1$. The angle between $l_1$ and $x$−axis, $\beta_1 = atan2((y_1 - y_0), (x_1 - x_0))$, is the actual direction of the robot displacement between time $T_0$ and $T_1$.

The absolute intersection angle between $l_0$ and $l_1$, $\delta$, is the deviation between the actual direction of the robot locomotion and the target direction. It can be calculated as:

$$\delta_{(\beta_0, \beta_1)} = \begin{cases} 2\pi - |\beta_1 - \beta_0| & (|\beta_1 - \beta_0| > \pi) \\ |\beta_1 - \beta_0| & (|\beta_1 - \beta_0| \leq \pi) \end{cases} \quad (5)$$

Note that we pick the smallest angle between the two lines. To perform well for a directed locomotion task, $\delta_{(\beta_0, \beta_1)}$ should be as small as possible. However, just minimizing $\delta_{(\beta_0, \beta_1)}$ is not enough to achieve successful directed locomotion.

In addition to moving in the right direction, i.e., minimizing $\delta_{(\beta_0, \beta_1)}$, the robot should move as far as possible in the target direction. Therefore, we calculate distance travelled by the robot in the target direction by projecting the final position at time $T_1$, $(x_1, y_1)$, onto the target direction $l_0$, i.e., the point $p(x_p, y_p)$; we denote this point as $p = (x_p, y_p)$. The distance travelled is then

$$\mathcal{D}_{(p,p_0)} = \text{sign} \, |p - p_0|, \tag{6}$$

where $|p - p_0|$ is the Euclidean distance between $p$ and $p_0$, and sign $= 1$ if $\delta_{(\beta_0,\beta_1)} < \frac{\pi}{2}$ (noting that $\delta_{(\beta_0,\beta_1)}$ is an absolute value) and sign $= -1$ otherwise. The $\mathcal{D}_{(p,p_0)}$ is thus negative when the robot moves in the opposite direction. To further penalize deviating from the target direction we calculate the distance between $(x_1, y_1)$ and $(x_p, y_p)$ :

$$\mathcal{P}_{(p,p_1)} = \omega \, |p_1 - p|, \tag{7}$$

where $|p_1 - p|$ is the Euclidean distance between $p_1$ and its projection $p$ on the target direction line $l_0$. $\omega$ is a constant scalar penalty factor, further determining the relative importance of the deviation. In our experiments we use $\omega = 0.01$. A naive version of the fitness function would be:

$$\mathcal{F}_{(\mathcal{D},\mathcal{P},\delta)} = \frac{\mathcal{D}_{(p,p_0)}}{\delta_{(\beta_0,\beta_1)} + 1} - \mathcal{P}_{(p,p_1)}, \tag{8}$$

where $(\delta_{(\beta_0,\beta_1)} + 1)$ aims to guarantee that the denominator does not equal zero. Particularly, $\mathcal{F}_{(\mathcal{D},\mathcal{P},\delta)} = \mathcal{D}_{(p,p_0)}$ when the final position $p$ of a locomotion is exactly in the target direction line $l_0$, i.e., $\delta_{(\beta_0,\beta_1)} = 0$ and $\mathcal{P}_{(p,p_1)} = 0$.

While $\mathcal{F}_{(\mathcal{D},\mathcal{P},\delta)}$ is proportional to $\mathcal{D}_{(p,p_0)}$, and inversely proportional to deviation $\delta_{(\beta_0,\beta_1)}$ and penalty $\mathcal{P}_{(p,p_1)}$, this does not yet entirely express all desirable features of a good directed locomotion. Specifically, we not only care about the final position of the robot, but also about how is the trajectory of the locomotion from starting position to end position.

To illustrate this we consider the trajectories marked $Tra.1$ and $Tra.2$ in Figure 7. Although the robot has the same starting and end position for both trajectories, $Tra.1$ is a more efficient way of moving between the two points. Therefore, we would expect the controller of $Tra.1$ to have a higher fitness than that of $Tra.2$. In general, we aim to evolve controllers that modular robots move from start to finish as efficiently as possible, i.e., in a straight line. Putting this all together we obtain the following fitness function to measure the performance of controllers for directed locomotion:

$$\mathcal{F}_{(\mathcal{D},\mathcal{P},\delta,\mathcal{L})} = \frac{|\mathcal{D}_{(p,p_0)}|}{\mathcal{L} + \varepsilon} \left( \frac{\mathcal{D}_{(p,p_0)}}{\delta_{(\beta_0,\beta_1)} + 1} - \mathcal{P}_{(p,p_1)} \right), \tag{9}$$

where $\varepsilon$ is an infinitesimal constant, the length of the trajectory $\mathcal{L}$ is calculated by summing the distances between two neighbouring positions in the trajectory that consist of ten positions during an evaluation. This fitness function is proportional to $\mathcal{D}_{(p,p_0)}$, but inversely proportional to $\mathcal{L}$ and $\delta_{(\beta_0,\beta_1)}$. That is, the fitness function rewards higher speed in the target direction (as measured through $\mathcal{D}_{(p,p_0)}$) and penalizes the length of trajectory $\mathcal{L}$ and the deviation from the target direction $\delta_{(\beta_0,\beta_1)}$.

## 4 Experiments

Experimental work is carried out in simulation as well as on real hardware. The underlying logic is to execute the learning algorithms in simulation and to test the best learned controllers on physical robots afterwards.

### 4.1 Experimental Setup

**Simulation** This work uses our own custom framework, $Revolve^1$ [26], based on $Gazebo^2$, that implements the components for running the Triangle of Life experiments [26]. It enables us to test the parts of the system as well as to set an entire environment for the complete evolutionary process. All experiments were performed using an infinite plane environment to avoid any extra complexity.

To learn the controllers for directed locomotion, we run Bayesian optimization and HyperNEAT with 1500 fitness evaluations using a 60 seconds test period per evaluation for each robot and each target direction. We repeat each learning run 10 times with different random seeds to compensate for the stochastic nature of the learning methods and aggregate the results over these 10 runs. All together this means 2 learners × 9 robots × 5 target directions × 10 repetitions that took about eight weeks time on a Linux computer with a 32 cores, 3.8GHz CPU and 64GB RAM.

**Real-world** We construct three representative robots with different shapes from the test suite, *Spider9*, *Gecko7*, and *BabyA*, as shown in Figure 8. The compo-

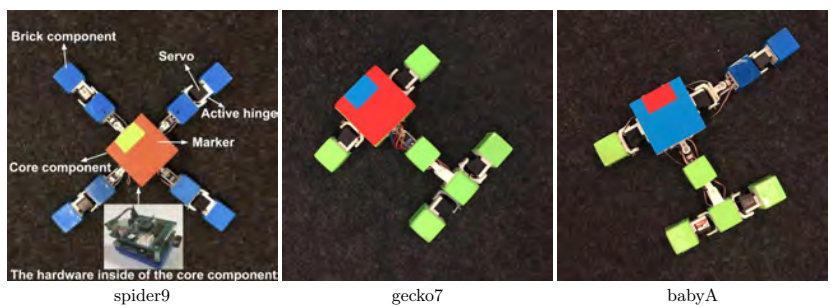

spider9      gecko7      babyA

Fig. 8: The prototypes of the three physical modular robots, Spider9, Gecko7, BabyA. The 3D printed components include *bricks*, *core*, *active hinges*. The electronic hardware inside of the core components includes Raspberry Pi with hat board design on top and battery mounted below. The combinations of the colorful papers on the top of robot heads are identification tags.

---

[1] https://github.com/ci-group/revolve
[2] http://gazebosim.org/

nents *bricks*, *core*, *active hinges* are printed in a 3D printer, and then assembled by hand with the electrical components including servos, Raspberry Pi microcomputer (with hat board), cables, and a battery. The combination of the colorful papers on the top of the robots' heads are the identification tags used to recognize the position of the robots by the overhead camera localization system.

In the real world experiments we do not replicate the learning processes. Instead, we validate the outcomes of learning: we run the best learned controllers on the robots and compare the real world performance with the simulated one. For this purpose, we test the best three controllers for each robot and each target direction and have the controller run for a period of 60 seconds in an arena of 4m×3m. The performance of the best learned controllers on real robots are evaluated with an overhead camera system above the arena that can recognize the positions of the robots on the performed locomotion trajectories. These positions are recorded over a 60 seconds test period and fed into the same fitness function as used in simulation. To cope with random effects we repeat each experiment three times. This implies 3 robots × 3 controllers × 5 directions × 3 repetitions: 135 real world experiments in total. All experiments in the real world took about two weeks time to perform because of the practicalities (assembly, servo breakdowns, system calibration, software errors).

### 4.2   Results with Simulated Robots

In this section, we present the experimental results from different perspectives. We show the development of the fitness over time during the learning process, but note that the numerical values come from Equation 9 and have no interpretable meaning. Therefore we also display the curves that show the deviation and locomotion speed and we plot the trajectories of robots using the best controllers.

The plots that show how fitness values change over time during the learning process are shown in Figure 9. Each sub-figure shows the average best fitness for a robot in five target directions, averaged over ten repetitions. The solid lines and dashed lines show the results for Bayesian optimization and HyperNEAT, respectively. These curves show a clear difference between the two learners. Bayesian optimization obtains higher fitness by the end of the learning period and learns much faster than HyperNEAT in the first hundreds of evaluations. It flattens out after about 500 evaluations and keeps growing at a slow pace. HyperNEAT performs a different learning processes at a more or less constant rate. Under the conditions of our experiments (1500 evaluations) it is inferior to Bayesian optimization considering the achieved fitness values at the end of the learning period. Specifically, we are concerned with morphologically evolving robot systems, where a newborn robot must acquire some basic behaviours after birth quickly. This implies that we are interested in the first part of the learning curves and not in the potential results after long times.

As noted above, the fitness values calculated by Equation 9 have no interpretable meaning. Therefore we also display the curves that show locomotion speed in Figure 10. These figures exhibit the same overall trends as Figure 9, but now with specific information about speeds. Bayesian optimization is faster in the beginning and for most robots it achieves higher speeds by the end as well. Regarding speed, our robots perfrom locomotion at 1 to 2 meters per minute, depending on the shape and the size. The fastest robot is gecko12 with speeds close to 2 meters/minute, while its smaller / larger versions (gecko7 / gecko17) only achieve about 1.3 meters/minute.

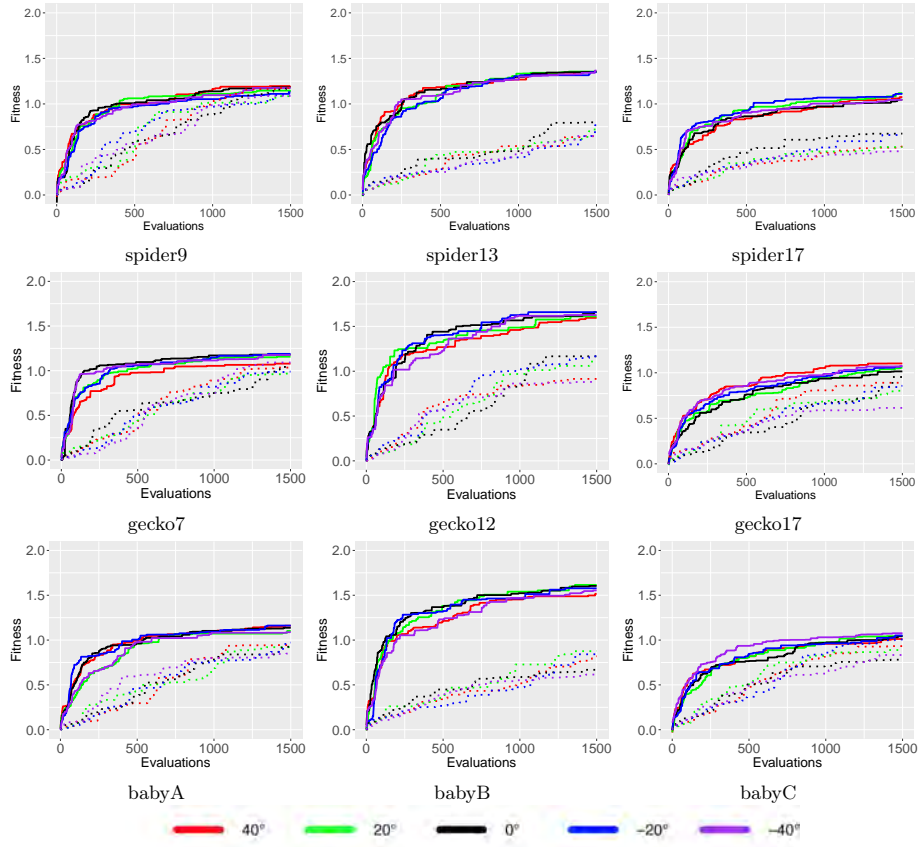

Fig. 9: The best fitness for Bayesian optimization (solid lines) and HyperNEAT (dashed lines) averaged over 10 runs. Colours represent target directions, red, green, black, blue and purple correspond to $40°, 20°, 0°, -20°$, and $-40°$, respectively. To keep figures uncluttered, we do not show the standard deviations.

Next we show the deviation $\delta_{(\beta_0,\beta_1)}$ of the locomotion with the best controllers in Figure 11. Considering that we are concerned with the task of directed locomotion, this is a a highly meaningful measure. We observe that for all robots and all directions, $\delta_{(\beta_0,\beta_1)}$ gradually decreases to low values close to zero. This indicates correctly directed locomotion towards the target directions. Just as before, we notice that Bayesian optimization performs better than HyperNEAT, although the differences are less prominent. The most apparent difference between the two learners is the consistency. Bayesian optimization quickly reaches and keeps low deviation values, while HyperNEAT experiences periods of deteriorating performance, where the deviations (temporarily) grow.

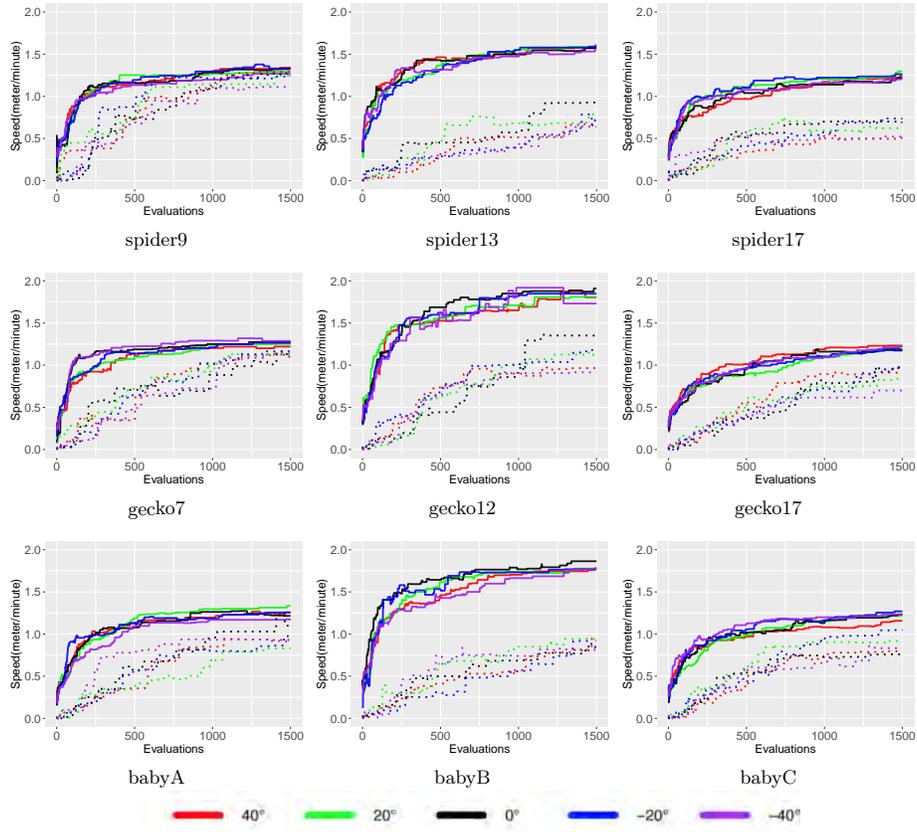

Fig. 10: Locomotion speed of the best controller for Bayesian optimization (solid lines) and HyperNEAT (dashed lines) averaged over 10 runs. Colours represent target directions, red, green, black, blue and purple correspond to $40°, 20°, 0°, -20°,$ and $-40°$, respectively. To keep figures uncluttered, we do not show the standard deviations.

Last but not least we inspect the trajectories of the robots with the best controllers. For each robot, each target direction, and both learners we select the top three controllers (with the highest fitness) from the 15000 controllers (1500 evaluations per run, 10 repetitions) that have been generated and tested by the given learner. Then we average out these three trajectories such that we obtain just one trajectory per robot, target direction, and learner, as shown in Figure 12.

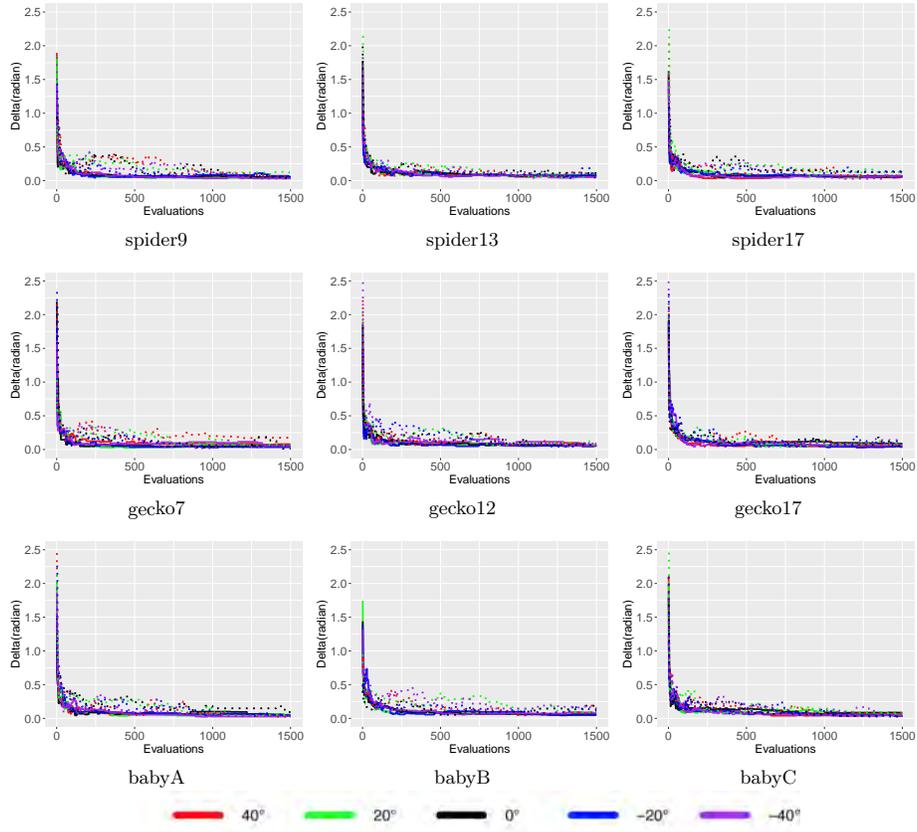

Fig. 11: Deviation from the target direction with the best controller for Bayesian optimization (solid lines) and HyperNEAT (dashed lines) averaged over 10 runs. Colours represent target directions, red, green, black, blue and purple correspond to $40°, 20°, 0°, -20°$, and $-40°$, respectively. To keep figures uncluttered, we do not show the standard deviations.

These trajectories support the observations based on the previous figures, the robots show adequate behaviour following the target directions. Overall, the solid lines tend to be longer and more accurate than the dashed ones. This confirms that Bayesian optimization is better than HyperNEAT. However, there are differences between the robots' behaviours. Maximizing the distance in the target direction, $\mathcal{D}_{(p,p_0)}$, is rewarded in the fitness function, as well as minimizing the deviation from the target directions and evolution can lead to different trade-offs between these two desirable properties. The trajectories of spider9 are almost exactly on the lines of the target directions. On the other hand, the solid red line of spider13 deviates far from the target direction, but scores a high fitness value because of the long distance it covers. The solid purple line of gecko12 shows a similar effect.

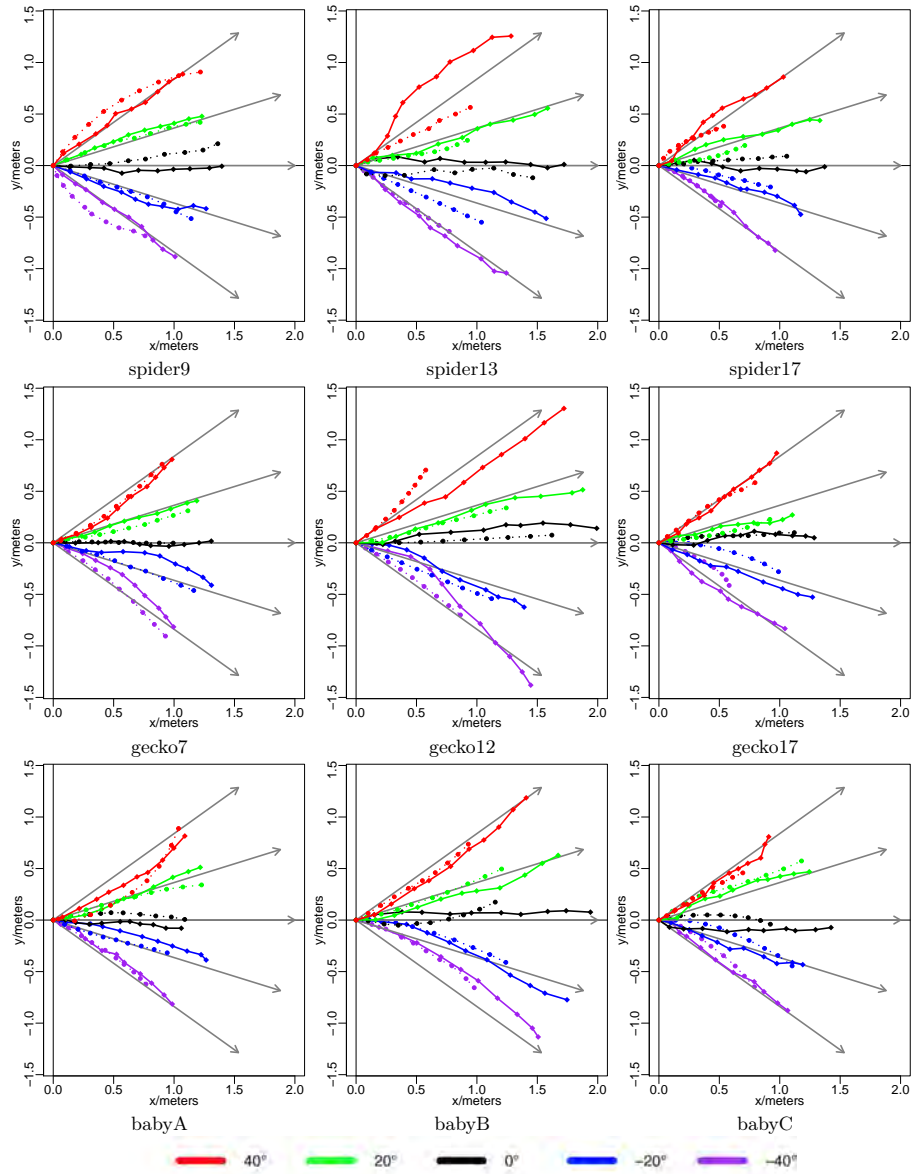

Fig. 12: Trajectories of the top three controllers averaged for each robot and each direction. for Bayesian optimization (solid lines) and HyperNEAT (dashed lines) averaged over 10 runs. Colours represent target directions, red, green, black, blue and purple correspond to $40°, 20°, 0°, -20°$, and $-40°$, respectively. The grey arrows show target directions.

In summary, the experiments in simulation show that successful controllers can be learned for directed locomotion on modular robots within a few hundreds of evaluations (learning trials) using Bayesian optimization as learning method. Figure 9 and Figure 10 indicate that 300 evaluations (learning trials) can already deliver decent performance. Spending a minute per learning trial this would cost about four hours. Whether or not this is fast enough depends on the specific circumstances, but in general, four hours to learn some desired behaviour on a 'newborn' robot does sound practicable.

### 4.3 Results with Real Robots

According to [40], less than 1% of evolutionary robotics studies tested the generated behaviours on real robots. However, due to the wear and tear of hardware and the infamous reality gap, the results obtained in simulations can be far off those observed on the actual hardware. Therefore, we perform additional experiments on three physical robots to see how the simulated behaviours hold up in the real world. For each real robot, the best three learned CPG controllers are evaluated with an overhead camera localization system. The resulting three fitness values for each real robot in each target direction are presented and compared with the simulated results in Figure 13. Each green bar shows the fitness of a top controller in simulation. Each orange bar shows the average fitness of three repetitions that a robot with the same top controller performed in the real-world. The three blue points in the column of an orange bar show the fitnesses of this controller for three repetitions in the real-world environment. Since we tested top three controllers for each target direction, Figure 13 therefore shows three green and orange bars for each target direction. The combination of a green bar, a orange bar and three blue points are the performance from the same controller in a target direction. As we can notice, most of the fitness values of the best controllers in simulation are higher than their fitness in the real-world scenario. That is, in our experiments we encounter the infamous reality gap for the directed locomotion of modular robots. Interestingly, three controllers obtained higher average fitness in real-world than the simulation, namely, the third controller of gecko7 in the target direction $0°$ and $-20°$, and the third controller of babyA in the target direction $-20°$, as shown in Figure 13. There are many factors that can lead to the locomotion of physical modular robots with higher fitness, We discuss one of the most likely factors in section 5.

The trajectories of the physical robots demonstrate how the best controllers perform in the real world. We take the average position of the locomotion over top three controllers with three repetitions as the trajectories that the best controllers performed on the physical modular robots in the target directions, as shown in Figure 14. The average trajectories of the locomotion with top three controllers for three repetitions that the best controllers performed on the physical modular robots in the target directions, are shown in Figure 14.

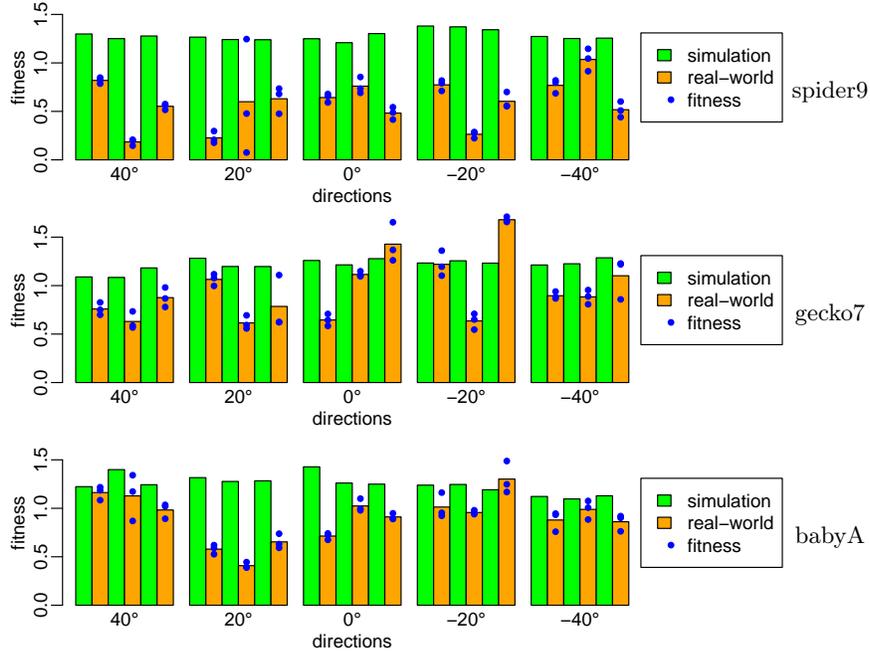

Fig. 13: The fitness comparison of the top three controllers in simulation and real world for the three robots (spider9, gecko7, and babyA) and each target direction. Each green bar shows the fitness of a top controller in simulation and its average fitness of three repetitions on the real robot is shown by the its right orange bar. The blue points in the column of an orange bar show the fitness values of a top controller for three repetitions in the real world. Note that some of the blue points coincide briefly because of their subequal fitness value.

The physical modular robots with the best controllers basically carry out well-directed locomotions in the target directions. For instance, let us consider the trajectories of spider9 that are almost on the lines of the target directions. While the second half of the average trajectory of spider9 in 40° is slightly deviated from the correct direction, the entire trajectory basically follows the direction in 40°. Comparing the average trajectories (solid lines) of top three controllers in Figure 12, it is apparent that directed locomotion is harder to be implemented in physical modular robots than in simulated robots. Particularly, the average trajectories of gecko7 in the target directions −20° and −40°, and babyA in the target directions −20° and −40° show the long locomotion distance and good directions (low $\delta_{(\beta_0, \beta_1)}$). Thus, they achieve high fitness that is close to the fitness in the simulation as shown in Figure 13. We conclude that the learned best controllers work well in the physical modular robots for directed locomotion.

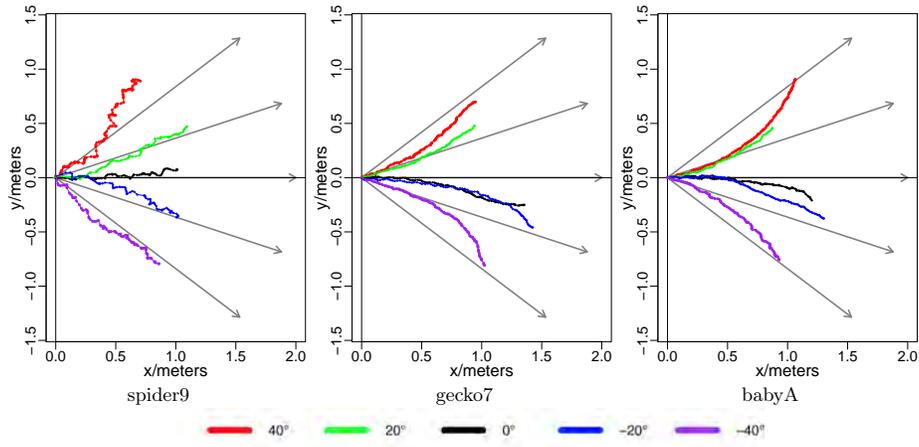

Fig. 14: The average trajectories of the top 3 controllers (with the highest fitness) for each robot and each direction in the real world. Each controller is run three repetitions and take the average trajectories. Five arrows in the sub-figures represent the five target directions. The trajectories are shown in five colours.

The video[3] of the physical modular robots shows the directed locomotion with the best learned controllers.

## 5 Discussion

### 5.1 Real World Behaviour

In the experiments with physical robots, we observed interesting behaviours that resemble animals in nature. The robot *spider9* performs the well-directed locomotion directly in the five directions without rotating. This is different from the trajectories of *gecko7* and *babyA* that perform curvilinear movement in target directions except the direction 0° rather than exactly follow the line in the target direction directly. Interestingly, animals like spiders and crabs also can go multi directions without rotating. The robot *spider9* performs biomimetic-like gait like zigzag for directed locomotion, as shown in Figure 14. Although the trajectories are averaged, the zigzag patterns are clear. We hypothesize that this is a gait characteristic for a spider topology.

For the robot *gecko7*, we observed that it achieves directed locomotion with curvilinear trajectories, as shown in the middle figure of Figure 14. That is, *gecko7* has to rotate for the locomotion in multiple directions except forward that results in a similar behavior to animals with similar morphologies, e.g., geckos and rabbits. Furthermore, we notice that the robot *gecko7* moves towards the target directions by pushing mainly from the combination of back legs and a waist. The front legs are mainly used to control the direction. Interestingly, animals like geckos and rabbits mainly use the back legs to drive the movement and control the direction. Nevertheless, they basically use their front limbs to control the locomotion direction rather than support the power to drive the locomotion.

---

[3] https://www.youtube.com/watch?v=Dhg1e8fqBgU

The robot *babyA* is generated by recombination from *spider9* and *gecko7*. As we can see in Figure 8, its topology is more similar to *gecko7* than to *spider9*. As a result, *babyA* performs quite similarly to *gecko7* in terms of trajectories, as shown in Figure 14. Moreover, the real *babyA* achieves directed locomotion with curvilinear trajectories that are similar to the trajectories of *gecko7*, but quite different from the trajectories of *spider9*. In addition, *babyA* has not performed the zigzag locomotion as the robot *spider9*.

The robustness of learned controllers in the real world is always an important issue. We therefore test this by answering the following question:

> How does a controller, learned for target direction $A_1$, perform in another target direction $A_2$ that it was not previously learned for?

We take the trajectories of a controller, originally obtained for target direction $A_1$, and calculate the fitnesses of these trajectories using target direction $A_2$. By this way, we evaluate the robustness of the controllers using other four target directions. The resulting robustness are shown in Table 5. The fitnesses in the diagonals of three robots represent the performance of the controllers in the original target directions. As we can see, although the controllers perform the best fitness in the original target directions (diagonals), they generally work well for the close target directions. For instance, the controller of *babyA* in a target direction $0°$ with a fitness of 0.88 achieves the fitnesses of 0.58 and 0.74 in the close target directions $20°$ and $-20°$, respectively. While the fitness values of a controller in other target directions are lower than it in original direction, it still obtains good fitness values in the close target directions. Therefore, we include that the CPG controllers are robust for the nearby target directions.

### 5.2 Reality Gap

The mismatch between simulation studies and experiments in real environments is always an important aspect of any research. As we highlighted in the experiments, in this work we notice the reality gap too. As exhibited in Figure 13, most of the fitness values for the physical modular robots are lower than the fitness in simulation. There are multiple possible causes of the reality gap, these are discussed in the following.

**Inconsistent friction**: It is demanding to design the same friction for the experiments in simulation and real-world because the friction in real world is difficult to be measured. Moreover, the friction between the physical robots and floor is dynamic when the robots is moving. In the physical experiments, we notice that the locomotion with the same controller becomes different when the components of the physical robots slide or get stuck on the floor. This is one of the main factors that lead to the reality gap, and is difficult to be solved.

| testing in→ | | 40° | 20° | 0° | −20° | −40° |
|---|---|---|---|---|---|---|
| spider9 | 40° | 0.52 | 0.44 | 0.27 | 0.10 | -0.01 |
| | 20° | 0.47 | 0.49 | 0.35 | 0.19 | 0.07 |
| | 0° | 0.25 | 0.47 | 0.63 | 0.46 | 0.24 |
| | −20° | 0.05 | 0.17 | 0.35 | 0.55 | 0.54 |
| | −40° | 0.0 | 0.05 | 0.21 | 0.47 | 0.77 |
| gecko7 | 40° | 0.75 | 0.69 | 0.40 | 0.17 | 0.04 |
| | 20° | 0.72 | 0.82 | 0.59 | 0.30 | 0.10 |
| | 0° | 0.31 | 0.65 | 1.06 | 0.96 | 0.56 |
| | −20° | 0.16 | 0.45 | 0.86 | 1.18 | 0.98 |
| | −40° | 0.03 | 0.15 | 0.40 | 0.74 | 0.96 |
| babyA | 40° | 1.09 | 0.75 | 0.39 | 0.14 | 0.02 |
| | 20° | 0.54 | 0.55 | 0.33 | 0.15 | 0.04 |
| | 0° | 0.29 | 0.58 | 0.88 | 0.74 | 0.42 |
| | −20° | 0.18 | 0.47 | 0.85 | 1.09 | 0.71 |
| | −40° | 0.02 | 0.11 | 0.33 | 0.65 | 0.91 |

Table 5: Robustness of the top learned controllers for different directions in the real world. The data shows the fitness that the top controllers of three robots in 5 directions (the second column) perform in another directions (the first row). The fitness value in a cell is averaged over three controller and three repetitions for each controller. The dark and light color represent the high and low fitness value respectively.

**Assembly errors**: The modular robots from simulation to real-world need to be assembled by hand. Thus, it is inevitable that the manual assembling leads to some errors. There are three main issues that are associated with manual assembling. First, the physical gap in the joints of the modular robots cause the different actuation. In this case, the physical modular robot perform a different locomotion from the robot with the same controller in the simulation. Second, the manual assembling is difficult to implement the physical modular robots with the exactly same size as in simulation. In particular, the position errors of the components in vertical usually cause the fatal errors. For instance, a "leg" of the physical robots is 0.1 cm off the floor since the assembly errors, which probably causes the "leg" to slide and further changes the locomotion. Third, the assembly gaps causes the physical robots have the model mismatch from the simulated modular robots. Although the model mismatch is in practice not too large and the learned CPG controller performs good robustness, a slight discrepancy may lead to a serious error, particularly in the task of directed locomotion. These errors can be accumulated and have a big impact on the behavior of directed locomotion.

**Physical weights**: The physical modular robots consist of many hardware including 3D printed components, servos, cables, microcomputer, and a battery. Correctly modeling these physical components is difficult. Therefore, the modular robots in simulation and real-world inevitably have different weights. The different weights impact the locomotion pattern, particularly the robots *gecko7* and *babyA*. Moreover, the different weight is a factor that cause the inconsistent friction.

**Initial directions**: At the start of each test, we position the physical robots in the correct direction $0°$ manually. Because of this, the deviations of the actual positioned direction from the direction $0°$ are inevitable. This usually causes that the same controller performs the locomotions with the different actual directions. Unlike most factors that cause a reality gap with a decreased fitness value, the initial direction errors probably decrease or increase the fitness value. For instance, the locomotion with the best controller deviates the target direction a little in the simulation, the initialization errors may cause the physical robot with the same controller performing a smaller or bigger deviation.

**Battery power**: The battery power decreases over time. We notice that the physical modular robots move faster with the fully charged battery than in the low battery mode. Therefore, the physical modular robots driven by the same controller generally have better performance (higher fitness) with fully charged battery than the low battery. Although this is not so remarkable as the issues above, it can still be a cause of the reality gap.

**Camera localization**: The camera localization system on the top of the arena is developed to recognize the position of the physical robot during the locomotion. Thus, the deviations of the localization also cause the different fitness for the locomotions with the same controllers.

**Servo quality**: Even though we use the same servos, the different servos quality cause that the same controller leads to a different locomotion.

## 6 Concluding Remarks

In this paper, we address the problem of learning sensory-motor skills in morphologically evolvable robot systems where the body of newborn robots is a random combination of the bodies of the parents. In particular, we consider a vital task, directed locomotion. We compare the popular HyperNEAT algorithm with Bayesian optimization using a fitness function that balances the distance travelled in a desired direction and the deviation angle between the actually travelled and the desired directions. We test the generality and scalability of the methods on a test suite of nine robots with different shapes and sizes for five different target directions. The results indicate that both methods learn good controllers across all of these morphologies and target directions, but Bayesian optimization outperforms HyperNEAT in terms of higher fitness values and faster learning speed. We noted a sweat spot around 300 learning trials that seem to be sufficient to learn good values for 20-30 parameters of the controller and deliver good performance in most cases. This is a large improvement over earlier results and even feasible to do on real robots. Analyzing the results of the locomotion speeds and the deviation angles, we could establish that our fitness function provides effective guidance for the learning algorithm by adequately balancing the locomotion speed and the direction of movement. Inspecting the trajectories obtained by using the best learned controllers demonstrated that the robots performed directed locomotion successfully. Nevertheless, a comparison of the results in the simulated and real robots shows that our system suffers from the infamous reality gap.

Ongoing work is aiming at a threefold extension of this study. The first direction concerns the improvement of the learning algorithm. While Bayesian optimization is very data-efficient, it is not time-efficient since its computation time increases cubically over evaluations. In addition, the performance of Bayesian optimization grows slowly after the initial stage. Therefore, we are working on a combination of evolutionary algorithms with Bayesian optimization to achieve lower computation times and higher accuracy.

The second line of research is to change the currently used open loop robot controller that works without feedback from perception. The enhanced system will employ a closed loop controller with feedback on the actual orientation of the robot. This will extend the applicability beyond a lab setup and hopefully reduce the reality gap by correcting locomotion towards the target direction. Moreover, the robustness (see Table 5) of the controllers can be improved by the closed loop directed locomotion. In a closed loop system, the deviation of actual traveling direction can be corrected real time.

Finally, we are integrating infant learning (as investigated here) in the grand evolutionary loop of the Triangle of Life. The resulting system features a morphologically evolving robot population, where reproduction among two parents delivers a new robot that performs learning right after 'birth' so as to maximize the potential of its morphology. This represents a novel combination of evolution and learning with unprecedented opportunities for researching evolution in robots and investigating hypotheses about natural evolution as well.

## Acknowledgment

The authors would like to thank Maarten van Hooft for his valuable suggestions and helpful coding. Matteo De Carlo was funded by the ARE@VU project.